%% file: main.tex
\documentclass[sigconf]{acmart}

\usepackage{algorithm}
\usepackage{algorithmic}
\usepackage{multirow}
\usepackage{subfigure}

\AtBeginDocument{%
  \providecommand\BibTeX{{%
    \normalfont B\kern-0.5em{\scshape i\kern-0.25em b}\kern-0.8em\TeX}}}

\copyrightyear{2023} 
\acmYear{2023} 
\setcopyright{acmlicensed}\acmConference[MM '23]{Proceedings of the 31st ACM International Conference on Multimedia}{October 29-November  3, 2023}{Ottawa, ON, Canada}
\acmBooktitle{Proceedings of the 31st ACM International Conference on Multimedia (MM '23), October 29-November 3, 2023, Ottawa, ON, Canada}
\acmPrice{15.00}
\acmDOI{10.1145/3581783.3612178}
\acmISBN{979-8-4007-0108-5/23/10}

\newcommand{\nosection}[1]{\vspace{2pt}\noindent\textbf{#1.}}
\newcommand{\problem}{FL with non-IID data}
\newcommand{\modelname}{\textbf{FedRANE}}
\newcommand{\moduleA}{\texttt{LRA}}
\newcommand{\moduleB}{\texttt{GNE}}
\usepackage{graphics}
\usepackage{multicol}
\usepackage{multirow}
\usepackage{xcolor}
\usepackage{amsmath}
\usepackage{amsthm}
\usepackage{enumitem}
\usepackage{bm}
\usepackage{booktabs}
\usepackage{units}
\definecolor{orange}{RGB}{255, 127, 0}

\theoremstyle{definition}
\usepackage{subfigure}

\begin{document}

\title{Joint Local Relational Augmentation and Global Nash Equilibrium for Federated Learning with Non-IID Data}

\author{Xinting Liao}
\affiliation{
  College of Computer Science, Zhejiang University
  \country{China}\\
xintingliao@zju.edu.cn
}

\author{Chaochao Chen}
\authornote{Chaochao Chen is the corresponding author.}
\affiliation{
  College of Computer Science, Zhejiang University
  \country{China}\\
zjuccc@zju.edu.cn
}

\author{Weiming Liu}
\affiliation{
  College of Computer Science, Zhejiang University
  \country{China}\\
21831010@zju.edu.cn
}

\author{Pengyang Zhou}
\affiliation{
  College of Computer Science, Zhejiang University
  \country{China}\\
zhoupy@zju.edu.cn,
}

\author{Huabin Zhu}
\affiliation{
  College of Computer Science, Zhejiang University
  \country{China}\\
 zhb2000@zju.edu.cn,
}

\author{Shuheng Shen}
\affiliation{\institution{Tiansuan Lab, Ant Group \country{China}}
}
\email{shuheng.ssh@antgroup.com}

\author{Weiqiang Wang}
\affiliation{\institution{Tiansuan Lab, Ant Group \country{China}}
}
\email{weiqiang.wwq@antgroup.com}

\author{Mengling Hu}
\affiliation{
\institution{College of Computer Science, Zhejiang University \country{China}}
}
\email{humengling@zju.edu.cn}

\author{Yanchao Tan}
\affiliation{
\institution{College of Computer and Data Science, Fuzhou University \country{China}}
}
\email{yctan@fzu.edu.cn}

\author{Xiaolin Zheng}
\affiliation{
  College of Computer Science, Zhejiang University
  \country{China}\\
 xlzheng@zju.edu.cn
}

\renewcommand{\shortauthors}{Xinting Liao et al.}







\begin{abstract}
Federated learning (FL) is a distributed machine learning paradigm that needs collaboration between a server and a series of clients with decentralized data.
To make FL effective in real-world applications, existing work devotes to improving the modeling of decentralized non-IID data.
In non-IID settings, there are intra-client inconsistency that comes from the imbalanced data modeling, and
inter-client inconsistency among heterogeneous client distributions, which not only hinders sufficient representation of the minority data, but also brings discrepant model deviations.
However, previous work overlooks to tackle the above two coupling inconsistencies together.
%
In this work, we propose \modelname, which consists of two main modules, i.e., local relational augmentation (\moduleA) and global Nash equilibrium (\moduleB), to resolve intra- and inter-client inconsistency simultaneously.
Specifically, in each client, \moduleA~mines the similarity relations among different data samples and enhances the minority sample representations with their neighbors using attentive message passing.
In server, \moduleB~reaches an agreement among inconsistent and discrepant model deviations from clients to server, which encourages the global model to update in the direction of global optimum without breaking down the clients' optimization toward their local optimums.
We conduct extensive experiments on four benchmark datasets to show the superiority of \modelname~in enhancing the performance of \problem.
\end{abstract}



\maketitle

\setlength{\floatsep}{4pt plus 4pt minus 1pt}
\setlength{\textfloatsep}{4pt plus 2pt minus 2pt}
\setlength{\intextsep}{4pt plus 2pt minus 2pt}
\setlength{\dbltextfloatsep}{3pt plus 2pt minus 1pt}
\setlength{\dblfloatsep}{3pt plus 2pt minus 1pt}
\setlength{\abovecaptionskip}{3pt}
\setlength{\belowcaptionskip}{2pt}
\setlength{\abovedisplayskip}{2pt plus 1pt minus 1pt}
\setlength{\belowdisplayskip}{2pt plus 1pt minus 1pt}

\input{./chapter/introduction.tex}

\input{./chapter/related.tex}

\input{./chapter/model.tex}

\input{./chapter/experiment.tex}

\vspace{-0.1cm}

\section{Conclusion}
In this work, we address the intra- and inter-client inconsistency of federated learning (FL) with non-IID data simultaneously.
Both intra- and inter-client inconsistencies together impact the performance of FL modeling, which causes an insufficient representation of imbalanced local data, and discrepant model deviations from clients to server.
To mitigate it, we propose \modelname, a federated learning framework with local relational augmentation (\moduleA) and global Nash equilibrium (\moduleB).
Specifically, \moduleA~tackles intra-inconsistency comes from imbalanced data, which mines the similarity
relations among different data samples and enhances the minority
sample representations with their neighbors.
GNE aims to handle inter-inconsistency among heterogeneous client distributions by reaching an agreement among 
discrepant model deviations, which improves both the global and local model performance.
We take extensive experiments on four benchmark
datasets to validate the effectiveness of \modelname.


\begin{acks}
This work was supported in part by the National Key
R\&D Program of China (No. 2022YFF0902704), the Ten
Thousand Talents Program of Zhejiang Province for Leading
Experts (No. 2021R52001), and Ant Group.
\end{acks}

\bibliographystyle{ACM-Reference-Format}
\balance
\bibliography{main}

\clearpage

\end{document}

%% file: chapter/introduction.tex
 \section{Introduction}
 \begin{figure}[t]
\centering
\includegraphics[width=\linewidth]{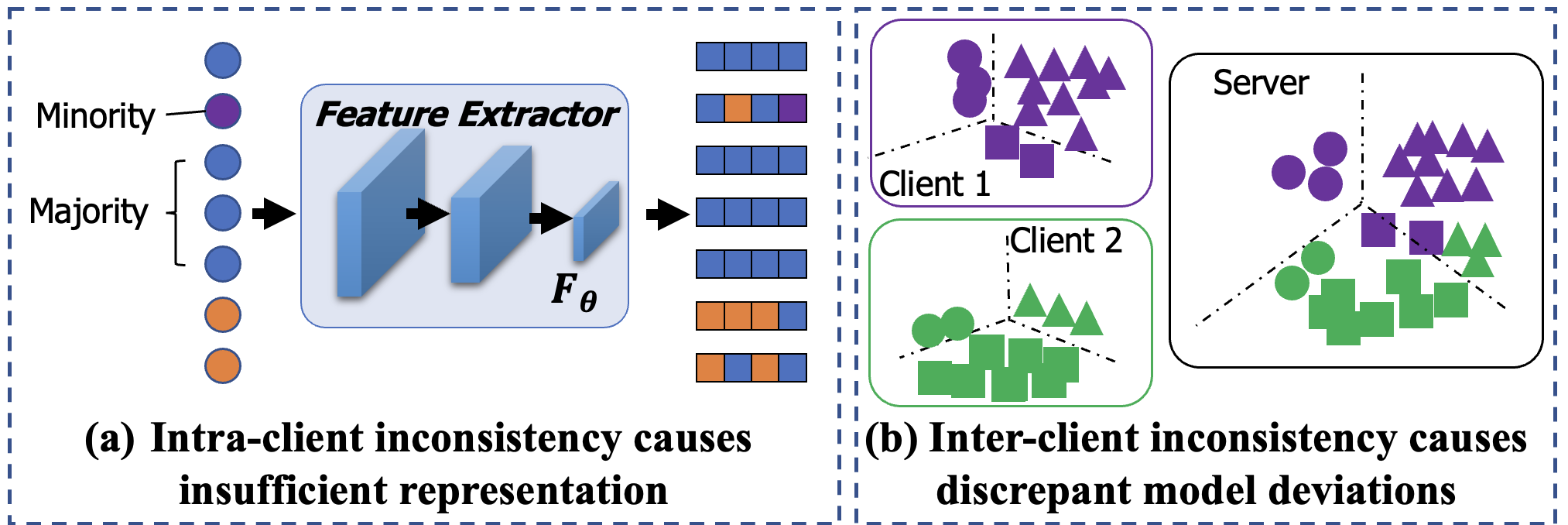} 
\caption{Motivation of \modelname.}
\vspace{-0.4cm}
\label{fig:motivation}

\end{figure}
%
Federated learning (FL) is a distributed machine learning paradigm, which consists of a server and a series of local clients~\cite{mcmahan2017communication,shen2019faster}.
With these collaborations between server and clients, the decentralized clients can enhance their model performance while keeping their data not exchanged with each other.
This provides a promising resolution to enhance the development of neural network modeling and preserve data privacy.
%
In many practical application settings, decentralized data have non-independent and  identical distributions (non-IID), which mainly challenges the development of FL~\cite{fallah2020personalized,oh2021fedbabu,dong2022spherefed}.
Since clients have to model their data to the local optimums that are inconsistent in non-IID settings, 
it is non-trivial to 
seek a consistent global optimum by aggregation~\cite{li2019convergence,karimireddy2019scaffold}.
%
%
%

In recent days, there are mainly three categories of efforts paid on \problem, i.e., (1)
improving the general global model performance, (2) enhancing the personalization of local model, and (3) achieving unified representation and personalized prediction simultaneously.
The \textit{first category} of work focuses on correcting the global models with regularization, e.g., FedProx~\cite{li2020federated}, controlling variance, e.g., SCAFFOLD~\cite{karimireddy2019scaffold}, and updating with momentum, e.g., SlowMo~\cite{wang2019slowmo}.
%
%
While the \textit{second category} of approaches prefer to encouraging diversification among clients, e.g., pFedMe~\cite{t2020personalized}.
%
%
The \textit{last category} of methods, e.g., FedBABU~\cite{oh2021fedbabu}, decouple client model into two parts.
%
Thus one part is able to enhance the global performance by regularization, as the first category does, while the other part achieves personalization, similar with the second category.

%
However, most existing work overlooks two potential challenges that hinder the performance of \problem, due to the intra- and inter-client inconsistencies.
For one thing
, the intra-client inconsistency comes from the imbalanced data, i.e., the data amounts of different labels are diversifying in each client.
Current FL methods \textit{fail to sufficiently represent the minority of imbalanced data} (\textbf{CH 1}).
As depicted in Fig.~\ref{fig:motivation} (a), during modeling the imbalanced data locally, the model updating accounts more for the majority of data samples, while ignoring the minority of data, leading to insufficient and inaccurate representations~\cite{seidenschwarz2021learning}.
%
In this way, the predictor cannot reason about the correct labels corresponding to minority data samples, based on the ambiguous feature representations that are quite similar to the majority.
Unfortunately, current work makes no evident efforts to mitigate this challenge.

For another thing, inter-client inconsistency happens when each client individually achieves their local optimums.
The existing work \textit{overlooks to negotiate an agreement among inconsistent model deviations from clients to server} (\textbf{CH 2}).
Client 1 and client 2 in Fig.~\ref{fig:motivation} (b)
capture the local samples distributions to 
inconsistent representation spaces, and optimize towards discrepant directions inevitably.
%
%
Without handling such inter-inconsistency, the samples with the same class label are represented differently, even distinctively, among different clients, which leads to a less deterministic decision bound in server.
%
Several previous work tries to alleviate this inter-consistency, and minimizes the global empirical risk by (1) reducing variance~\cite{karimireddy2019scaffold}, (2) regularizing local optimization~\cite{li2020federated,acar2020federated}, or (3) accounting momentum~\cite{xu2019hybridalpha}.
However, simply focusing on minimizing global empirical risk degrades the local personalized performance~\cite{chen2021bridging}, while blindly weighting on local optimums causes global performance shrinkage~\cite{li2021fedrs,wang2021federated,cui2021addressing}.
Thus, it is necessary to adequately account for both global and local model performance, via negotiating an agreement among inconsistent client deviations.

In this work, we propose a federated learning framework with local relational augmentation and global Nash equilibrium (\modelname), to tackle intra- and inter-client inconsistencies, simultaneously.
For handling \textbf{CH 1} with intra-client inconsistency, we devise \textbf{local relational augmentation} (\moduleA) module in each client, which enhances sample representation with its neighbors, i.e., samples with high similarity.
\moduleA~first computes the similarity among a batch of data samples, and finds the neighbors of data samples based on the similarity.
Then \moduleA~enhances the data feature representation via attentive message passing among the neighbors of data samples. 
Besides, \moduleA~conducts contrastive 
discrimination to maintain the representations correspondence before and after augmentation, for the same sample.
Aiming at \textbf{CH 2} caused by inter-client inconsistency, we utilize \textbf{global Nash equilibrium} (\moduleB) module in server, which obtains an agreement among inconsistent deviations from clients to server.
Specifically, \moduleB~collects the updating deviations from different clients to server.
Then \moduleB~not only seeks a global optimization direction that maximizes the consistency among discrepant local model deviations, but also maintains clients' optimizations towards their local optimums. 
This can be formulated as a Nash bargain problem~\cite{nash1953two}.
That is, the clients are players with inconsistent optimization objectives, and they seek a Pareto optimal solution, i.e., a solution where any modification will have a negative average relative change, to collaborate and maximize the overall effectiveness.
Next, \moduleB~optimizes for Pareto optimal solution with a multi-task optimization algorithm efficiently, and aggregates client models with the final Pareto optimal solution.
%

To conclude, we are the first, as far as we know, to address both the intra- and inter-client inconsistencies of \problem~simultaneously.
The main contributions are :
(1)
We enhance the discrimination of the minority sample representations with its related neighbors, which mitigates the intra-client inconsistency during local modeling.
(2) We optimize the combination of different client deviations to a consistent updating direction in server, which not only minimizes the impact of the inconsistent clients deviations in federated aggregation, but also keeps the clients' optimization towards their optimums unchanged.
%
(3) We conduct empirical studies on four benchmark datasets to prove the superiority of \modelname, 
compared with the state-of-the-art (SOTA) FL methods.

%% file: chapter/related.tex
\section{Related Work}
\subsection{Federated Learning with Non-IID Data}
In terms of the goal of optimization, we categorize the existing work related to \problem~as bellow:
%
(1) \textit{Global performance}, which focuses on correcting the global models to be well-performed with regularization, e.g., FedProx~\cite{li2020federated}, controlling variance, e.g., SCAFFOLD~\cite{karimireddy2019scaffold,liang2019variance}, and updating with momentum, e.g., SlowMo~\cite{wang2019slowmo,wang2021local}.
%
%
(2) \textit{Local performance}, which enhances a personalized model for each individual client
via utilizing the generalization capability of meta-learning~\cite{fallah2020personalized}, transfer learning\cite{luo2022disentangled}, knowledge distillation~\cite{huang2022learn,xie2023perada}, and so on.
For example, DFL~\cite{luo2022disentangled} utilizes  transferring learning to enhance the diversity of representation with task-correlated domain-specific attributes.
%
%
%
However, due to the coupling impact of intra- and inter-client inconsistency, simply enhancing the global performance will degrade the local performance, and vise visa \cite{chen2021bridging,wang2021federated,cui2021addressing}.
%
(3) \textit{Global and local performance}, which decomposes the neural network model in FL~\cite{liao2023hyperfed}, and separately improves global and local performance as the above two categories do. 
 Fed-RoD~\cite{chen2021bridging} consists of two classifiers to maintain the local and global performance, respectively.
 %
 %
FedBABU~\cite{oh2021fedbabu} and SphereFed~\cite{dong2022spherefed}  fix the classifier during training FL, and aggregate models following FedAvg~\cite{mcmahan2017communication}.
%
Though the decomposition approach disentangles the impact between global and local optimization, they fail when non-IID is serious.
%
Differently, \modelname~devotes to negotiating an agreement that not only improves global performance but also maintains local optimization.
%
Besides, several work, e.g., Wang et al.~\cite{wang2021addressing} and CLIMB~\cite{shen2022agnostic}, studies tackling the FL with imbalance data problem, which mainly focuses on mitigating the significant mismatch between local and global imbalance.
However, these methods either leak privacy due to computing a ratio-loss~\cite{wang2021addressing} with auxiliary data sampled from clients, or relies on a hand-crafted tolerance parameter~\cite{shen2022agnostic} to constrain the training loss among clients.
%
%
%
In this paper, we propose \modelname~to directly refine the representation of minority samples with its intra-client neighbors, without regularization from other clients or server.

\subsection{Multi-task Learning}
Multi-task learning (MTL) simultaneously solves multiple related learning problems while sharing information among tasks~\cite{ruder2017overview,zhang2021survey}.
The most popular MTL objective is to minimize the average loss over all tasks,
ignoring inconsistent tasks~\cite{liu2021conflict}.
%
%
Several centralized machine learning work is devised to address this challenge by mitigating conflicting gradients among different tasks.
MGDA~\cite{sener2018multi} 
aims to balance conflicting tasks and achieve a Pareto optimal solution.
%
%
PCGrad~\cite{yu2020gradient} identifies the presence of conflicting gradients
and projects each task gradient onto the normal plane of others to minimize conflicts.
CAGrad~\cite{liu2021conflict} offers a more comprehensive approach.
%
Moreover, Nash-MTL~\cite{navon2022multi}~finds a Pareto optimal that is invariant to changes in loss scale and produces balanced solutions across the Pareto front.
%
In the context of FL, two aspects of work utilize MTL paradigm, i.e., personalized FL with MTL, e.g., FedMTL~\cite{smith2017federated}, and fair FL with MTL encourages clients to behave uniformly, e.g., FedMGDA+~\cite{hu2020fedmgda+} and FedFA~\cite{ijcai2021p223}.
Personalized FL with MTL cannot guarantee both optimal global and local performance when client deviations are heavily inconsistent.
And current work on fair FL with MTL mainly focuses on maintaining uniform local performance among clients.
Besides, they utilize MTL techniques derived from MGDA and CAGrad, resulting in imbalanced solutions
~\cite{navon2022multi}.
%
Differently, \modelname~
formulates aggregating  model with inconsistent deviations as a Nash bargaining problem, enjoying a balanced agreement that maximizes both global and local performance.

%% file: chapter/model.tex
\section{Method}
\begin{figure*}[t]
\centering
\includegraphics[width=0.98\linewidth]{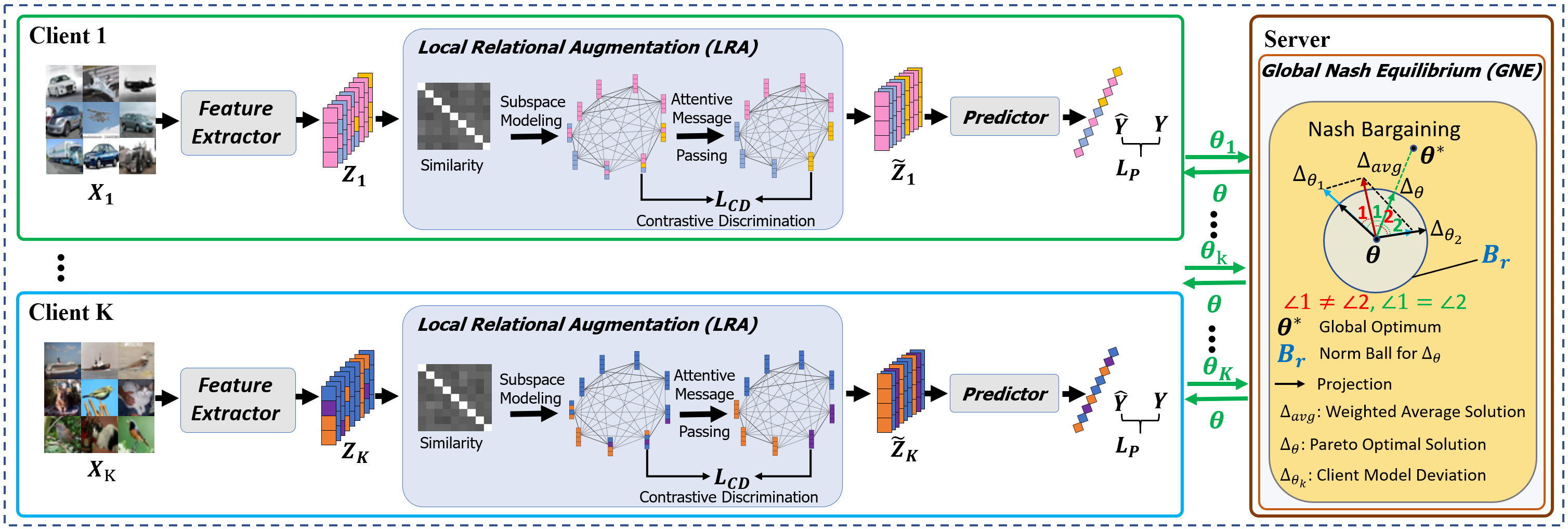} 
\caption{Framework of \modelname. Every client contains \moduleA~for addressing intra-client inconsistency. 
Conventionally, server updates global model with weighted average solution, i.e., $\Delta_{\text{avg}}$ denoted by the red arrow, which is inevitably inconsistent with the global optimum.
In contrast, \moduleB~seeks the Pareto optimal solution, i.e., $\Delta_{\boldsymbol{\theta}}$ denoted by the green arrow, which not only updates towards the global optimum but also maintains the consistency of updating each local model. 
}
\label{fig:model}
\end{figure*}

\subsection{Problem Statement}
We first describe the problem and assumptions in this section.
For \problem, we assume a dataset decentralizes among $K$ clients, i.e.,  $\mathcal{D} =\cup_{k\in [K]} \mathcal{D}_k$ 
, where the data distributions of different $\mathcal{D}_k$ are non-IID.
The clients model their datasets locally, while the server collaborates clients' models to update a consistent model globally. 
In detail, each client contains $N_k$ data samples, i.e.,
$\mathcal{D}_k=\{ \boldsymbol{x}_{k,i}, {y}_{k,i}\}_{i=1}^{N_k}$, where the number of samples corresponding to different ${y}_{k,i}$ is \textit{inconsistent} intra- and inter-client.
The overall objective of \problem~is defined as below: 
\begin{equation}
{\operatorname{argmin}}_{\boldsymbol{\theta}} \mathcal{L}(\boldsymbol{\theta}; \boldsymbol{p})=\Sigma_{k=1}^K p_k \mathbb{E}_{\boldsymbol{x}\sim \mathcal{D}_k}[\mathcal{L}_k(\boldsymbol{\theta}; \boldsymbol{x}, y)],
\label{eq:pfl_problem}
\end{equation}
where $\mathcal{L}_k (\cdot)$ is the model loss at client $k$, $\boldsymbol{p}=[p_1, \dots, p_K]$, and $p_k$ represents its weight ratio in aggregation. 
Conventionally, the existing methods assign $p_k$ with the ratio of local sample number to global sample number, i.e., $p_k= \nicefrac{|
\mathcal{D}_k|}{|\mathcal{D}|}$.  
%
However, this cannot tackle inconsistent model deviations from clients to server well.
Because the clients with more data will dominate the model aggregation and degrade the performance of other clients.
In this work, we seek $\boldsymbol{p}$ that reaches a consistent global updating direction among all clients, and maximizes both the global and local performance.

\subsection{Framework Overview}
To address \problem, we depict the framework overview of \modelname~in Fig.~\ref{fig:model}. 
%
%
%
Each client similarly contains a neural network model consisting of a feature extractor module, a \moduleA~module, and a predictor module. 
The server owns a \moduleB~module, which aggregates client models with an agreement.
We first introduce the local modeling at each client $k$, and illustrate the global model aggregation in server later on.
For a batch of data at client $k$, we input them to the feature extraction module and \moduleA~module sequentially. 
The feature extractor $\mathcal{F}_{\boldsymbol{\theta}_k}(\cdot): \mathcal{X}\rightarrow \mathbb{R}^d$ maps a batch of input data $\boldsymbol{X}_k$ into a $d$-dimensional vector $\boldsymbol{Z}_k=\mathcal{F}_{\boldsymbol{\theta}_k}(\boldsymbol{X}_k)$ as feature representations.
Then using $\boldsymbol{Z}_k$, the \moduleA~constructs the data graph, i.e., the Laplacian matrix $\boldsymbol{L}_k$, based on the similarity matrix of samples.
With graph structure, the \moduleA~module applies attentive message passing to enhance the feature representation of each sample node with its neighbors, i.e.,  $\widetilde{\boldsymbol{Z}}_k=\mathcal{G}_{\boldsymbol{\theta}_k}(\boldsymbol{Z}_k, \boldsymbol{L}_k)$. 
Additionally, \moduleA~regularizes the representations of the same sample, before and after augmentation.
Finally, the predictor module, i.e., a multi-perception layer module, infers the sample label based on the augmented sample representation.
After local training, each client $k$ uploads its model parameters $\boldsymbol{\theta}_k$ to the server.
For global aggregation in server, \moduleB~addresses the inconsistent model updating deviations from clients to server.
\moduleB~first formulates the aggregation among different client models as a Nash bargaining problem, i.e.,  negotiating an agreement among inconsistent model deviations, and resolves it into a Pareto optimal solution via multi-task learning.
Then server sends the new and consistently updated global model back to clients.
This communication between server and clients iterates until \modelname~converges.

\subsection{\moduleA: Addressing Intra-client Inconsistency}
\nosection{Motivation}
 In this section, we introduce \moduleA~that address intra-client inconsistency, via obtaining distinguishable and sufficient representations for data samples.
\moduleA~explores the overall relational structure of representations in a batch, and augments the representations using message passing.
However, the representation relations are always sparse and undiscovered, which cannot be directly applicable and reliable for subsequent relational augmentation.
Meanwhile, message passing is supposed to bring refined and sufficient representations, while avoids the representation of minority samples  contaminated by that of the majority ones.
To mitigate it, \moduleA~first conducts sample relational mining with subspace modeling,
which uncovers the sparse and undiscovered relations of data representations via subspace modeling on a similarity matrix, e.g., Pearson correlation matrix.
%
Then \moduleA~refines sufficient sample representation via attentive message passing among its neighbors.
%
To avoid unexpected representation contamination, 
 \moduleA~further constrains the representation correspondence before and after message passing of the same data sample, using contrastive discrimination.
After that, the predictor feedback, i.e., the prediction loss, corrects \moduleA~to augments distinguishable and sufficient data representations corresponding to the ground truth class labels.

\nosection{Sample Relational Mining with Subspace Modeling}
%
To enhance the model's perception of minorities, \moduleA~discovers the graph structure of data in a batch and augments the data representations with their neighbors. 
For feature representations obtained from the feature extractor module, i.e., $\boldsymbol{Z}=\mathcal{F}_{\boldsymbol{\theta}}(\boldsymbol{X})$, we construct a graph to find their neighbors based on feature similarity~\cite{weimingSigir22}.
However, the data samples of different classes are imbalanced, which causes the relations among data samples to be undiscovered and sparse.
Motivated by Sparse LInear Methods (SLIM)~\cite{cheng2014lorslim,ning2011slim,liu2022collaborative}~which effectively mine the sparse and low-rank item-item relation in recommender systems, we mine the relations of minibatch sample representations via modeling the subspace weights of statistics similarity matrix.

%

Next, \moduleA~adopts Pearson correlation matrix~\cite{cao2022pkd}, i.e., $\boldsymbol{P}$, as the input of SLIM methods to mine the sample representation relations, i.e., the relational weights matrix $\boldsymbol{B}$, by the optimization objective:
\begin{equation}
\begin{array}{cc}
\min _{\boldsymbol{B}}  \frac{1}{2} \|\boldsymbol{P}-\boldsymbol{P} \boldsymbol{B}\|_F^2+\lambda_R \cdot\|\boldsymbol{B}\|_*^2 \quad
\text { s.t. }  \operatorname{diag}(\boldsymbol{B})=0,
\end{array}
\label{eq:slim}
\end{equation}
where $\|\cdot\|_F$ is the Frobenius norm, $\operatorname{diag}(\boldsymbol{B})=0$ penalizes trivial solution, $\lambda_R$ is the hyper-parameter, and $||\boldsymbol{B}||_{*} = \operatorname{Tr}({(\boldsymbol{B}^\top  \boldsymbol{B})}^{\frac{1}{2}})$ is the nuclear norm to attain low-rank matrix that enhances robustness and generalizations~\cite{cheng2014lorslim}. 
We optimize Eq.~\eqref{eq:slim} by minimizing its Lagrangian formulation as below:
\begin{equation}
\begin{array}{cc}
\min _{\boldsymbol{B}} \frac{1}{2} \|\boldsymbol{P}-\boldsymbol{P} \boldsymbol{B}\|_F^2+\lambda_R \operatorname{Tr}(\boldsymbol{B}^\top \boldsymbol{\Phi} \boldsymbol{B}) \quad \text { s.t. }  \operatorname{diag}(\boldsymbol{B})=0,
\end{array}
\label{eq:slim_l}
\end{equation}
where we denote $\boldsymbol{\Phi}= {(\boldsymbol{B}\boldsymbol{B}^\top)}^{-\frac{1}{2}}$.

We alternatively update $\boldsymbol{B}$ and $\boldsymbol{\Phi}$ to obtain the closed form of $\boldsymbol{B}$:
\begin{equation}
\boldsymbol{B}_{i, j}= \begin{cases}0, & \text { if } i=j \\ -\frac{{\boldsymbol{H}}_{i j}}{{\boldsymbol{H}}_{j j}}, & \text { otherwise. }\end{cases}
\end{equation}
We first treat $\boldsymbol{\Phi}$ as constant, and $\boldsymbol{H} = (\boldsymbol{P}^\top \boldsymbol{P}+ \lambda_R (\boldsymbol{\Phi}+ \boldsymbol{\Phi}^\top))^{-1}$.
Then taking $\boldsymbol{B}$ as a constant, we update $\boldsymbol{\Phi}$ with the constraint  $\boldsymbol{\Phi}= {(\boldsymbol{B}\boldsymbol{B}^\top)}^{-\frac{1}{2}}$.
We iterate this alternative updating until it converges, which finally captures an asymmetric matrix $\boldsymbol{B}$ with unknown positive definiteness.
To mitigate it, we finally build up the sample graph with adjacent matrix $\boldsymbol{A} = \nicefrac{(|\boldsymbol{B}|+ |\boldsymbol{B}|^\top)}{2}$, and Laplace matrix $\boldsymbol{L} = \boldsymbol{D} - \boldsymbol{A}$, where $\boldsymbol{D}$ denotes the degree matrix on graph.
Thus, \moduleA~obtains a graph $\boldsymbol{G} =(\boldsymbol{V},\boldsymbol{A})$, where $\boldsymbol{V}$ are nodes corresponding to every data sample, and $\boldsymbol{A}$ are edges connecting to them.
%

\nosection{Sufficient Sample Representation via Attentive Message Passing}
Next, \moduleA~enhances the data representation of each sample node by attentive message passing among their neighbors in a batch.
We start at $\boldsymbol{h}_{i}^0= \boldsymbol{z}_{i}$, 
and obtain attention weighted messages for node $i$ from its neighbors $\mathcal{N}_{i}$ in step $l$, i.e.,
\begin{equation}
\boldsymbol{h}_{i}^{l+1}=\sum_{j \in \mathcal{N}_{i}} \alpha_{i j}^l \boldsymbol{W}^l \boldsymbol{h}_{j}^l, 
\label{eq:mpn}
\end{equation}
where $\boldsymbol{W}^l$ represents the corresponding weight matrix of message passing step $l$, and $\alpha_{ij}$ is attention weight.
We compute the dot-product self-attention weight for each step $l$ as below:
\begin{equation}
\alpha_{i j}^l=\nicefrac{\boldsymbol{W}_m^l \boldsymbol{h}_{i}^l\left(\boldsymbol{W}_n^l \boldsymbol{h}_{j}^l\right)^T}{\sqrt{d}},
\label{eq:attention}
\end{equation}
where $d$ is the feature dimension, $\boldsymbol{W}_m^l$ and $\boldsymbol{W}_n^l$ are the weight matrix to receiving nodes and sending nodes, respectively.
We take $L$ steps to obtain final relational augmented feature representation, i.e., $\tilde{\boldsymbol{z}}_{i} = \boldsymbol{h}^L_{i}$.
With the attentive message passing among batch sample graph, \moduleA~captures the structural information to enhance feature representations of data samples, which alleviates the insufficient representation in modeling imbalanced data.

\nosection{Representation Correspondence for the Same Sample using Contrastive Discrimination}
We devise additional guidance signals via contrastive discrimination (CD), in order to maintain representation correspondence before- and after relational augmentation.
Without this correspondence constraint, message passing in Eq.~\eqref{eq:mpn} will inevitably contaminate the representations of minority samples by that of the majority ones, and fail to guarantee correct representations for prediction.
%
In detail, CD derives from contrastive loss, i.e., SimCLR loss~\cite{chen2020simple,liu2023joint,liu2021leveraging}, and encourages the different views of the same data sample to share similar class assignment distribution.
Given feature representations before and after augmentation, i.e., $\boldsymbol{Z}$ and $\widetilde{\boldsymbol{Z}}$, and batch size $B$, we 
concatenate them as $\hat{\boldsymbol{Z}} = [\boldsymbol{Z};\widetilde{\boldsymbol{Z}}]$.
Then take $\hat{\boldsymbol{z}}_{i}$ and $\hat{\boldsymbol{z}}_{B+i}$
(corresponding to $\boldsymbol{z}_{i}$ and $\tilde{\boldsymbol{z}}_{i}$) as the positive pair, and the remaining $2(B-1)$ sample pairs in a batch as negative, and compute the loss function as below:
\begin{equation}
\small
\mathcal{L}_{CD}= - \frac{1}{B} \sum^{B}_{i=1} \log \frac{\exp \left(\operatorname{sim}\left(\hat{\boldsymbol{z}}_{i}, \hat{\boldsymbol{z}}_{B+i}\right) / \tau_1\right)}{\sum_{j=1,j \neq i}^{2B} \exp \left(\operatorname{sim}\left(\hat{\boldsymbol{z}}_{i}, \hat{\boldsymbol{z}}_{j}\right) /\tau_1\right) } ,
\label{eq:L_cd}
\end{equation}
where $\tau_1 (\tau_1 \in[0,1])$ is temperature hyperparameter.
We also take Pearson coefficient as similarity.
Eq.~\eqref{eq:L_cd} encourages 
sample representations, before and after augmentation, to get consistent labels.

\nosection{Prediction and Optimization}
Given the relational augmented representation $\widetilde{\boldsymbol{Z}}$ as input, the predictor outputs its inference $\hat{\boldsymbol{y}}$.
The predictor is trained to minimize cross entropy $\operatorname{F}_{\text{ce}}(\cdot)$ as below: 
\begin{equation}
    \label{eq:pred} 
    \mathcal{L}_{\text{pred}} = \operatorname{F}_{\text{ce}} (\hat{\boldsymbol{y}}, \boldsymbol{y}).
\end{equation}
The overall local optimization objective is to minimize:
\begin{equation}
    \mathcal{L}=\mathcal{L}_{\text{pred}}+\lambda_{\text{CD}}  \mathcal{L}_{CD},
    \label{eq:loss}
\end{equation}
where $\lambda_{\text{CD}}$ is the hyperparameter.
In the end, we capture sufficient feature representation of data samples with \moduleA, and tackle the intra-client inconsistency due to imbalanced data.

\subsection{\moduleB: Tackling Inter-client Inconsistency}
\nosection{Motivation}
In this section, we provide the details related to \moduleB~that handles inter-client inconsistency, i.e., different clients individually model their own data to their local optimums without the knowledge of others.
To update the global model towards global optimum without breaking down the local optimization, 
\moduleB~in server requires negotiating an agreement with inconsistent deviations when aggregating client models.
As shown in Fig.~\ref{fig:model}, if the server inadequately accounts for these inconsistent deviations in aggregating client models, the updated global model direction, i.e., $\Delta_{\text{avg}}$ denoted by red arrow, will deviate from the global optimum~\cite{li2020federated}.
While simply regularizing the local model optimization with the constraints of global model will hurt the local model performance~\cite{cui2021addressing}.
\moduleB~trades off inconsistent model deviations from clients to server, and reaches a 
Pareto optimal solution, 
i.e., obtaining a global updating direction that is consistent with all client.
%
%
As shown in \moduleB~of Fig.~\ref{fig:model}, the Pareto optimal solution, i.e., $\Delta_{\boldsymbol{\theta}}$ denoted by the green arrow, is more balanced and shares the same angle with deviations of client 1 and client 2.
%
In detail, \moduleB~first collects the model deviations from clients to server, and formulates the combination of model deviations as a Nash Bargaining problem.
Then \moduleB~characterizes the Pareto optimal solution of this Nash bargaining problem, and approximates its value via an efficient multi-task optimization algorithm.
%

\nosection{Nash Bargaining Problem Formulation on Client Aggregation}
We formulate the aggregation as a Nash bargaining problem in the following. 
Specifically, \moduleB~first computes the different deviations from clients to server.
%
For a combination of the local model parameters in server, we can write it as: 
\begin{equation}
\label{eq:agg}
\boldsymbol{\theta}^{t+1} = \boldsymbol{\theta}^{t}+\Sigma_{k=1}^K p_k\left(\boldsymbol{\theta}_k^{t+1}-\boldsymbol{\theta}^{t}\right),
\end{equation}
where $\boldsymbol{\theta}^{t}$ is the global model, and $\boldsymbol{\theta}_k^{t}$ is the local model of the client $k$ at $t-$th communication.
Next, we denote the global updating direction as $\Delta_{\boldsymbol{\theta}}^{t+1}=\boldsymbol{\theta}^{t+1} - \boldsymbol{\theta}^{t}$ and the model deviation of client $k$ as $\Delta_{\boldsymbol{\theta}_{k}}^{t+1}=\boldsymbol{\theta}_k^{t+1}-\boldsymbol{\theta}^{t}$,
to rewrite Eq.~\eqref{eq:agg} as:
\begin{equation}
\label{eq:agg_grad}
    \Delta_{\boldsymbol{\theta}}^{t+1} = \Sigma_{k=1}^K p_k \Delta_{\boldsymbol{\theta}_{k}}^{t+1}.
\end{equation}

Since server collaborates discrepant client deviations to update a global model, i.e., Eq.~\eqref{eq:agg_grad}, we can formulate it as a
Nash bargaining problem, which balances inconsistent player utility functions and collaboratively maximizes the overall utility without hurting any player's utility.
%
%
Specifically, \moduleB~seeks an update vector $\boldsymbol{\Delta}_{\boldsymbol{\theta}}$ with the agreement set $B_{r}$, i.e., a ball of radius $r$ centered around zero, and a disagreement point at $0$, i.e., keeping current global model $\boldsymbol{\theta}$ unchanged.
We define the overall Nash bargaining problem as:
\begin{equation}
    \arg\max _{\boldsymbol{\Delta}_{\boldsymbol{\theta}} \in B_{r}} \Sigma_{k=1}^K \log [u_k(\Delta_{\boldsymbol{\theta}})],
\label{eq:nbp}
\end{equation}
where $u_k(\Delta_{\boldsymbol{\theta}})={\Delta_{\boldsymbol{\theta}_{k}}^\top \Delta_{\boldsymbol{\theta}}}$ 
 is the utility function of each client.
For all vectors $\Delta_{\boldsymbol{\theta}}$ such that $\forall_k:{\Delta_{\boldsymbol{\theta}_{k}}^\top \Delta_{\boldsymbol{\theta}}} >0$, 
the overall utility is monotonically increasing with the norm.
In this case, the unique optimal solution is exactly on the boundary of $B_r$, in terms of the Pareto optimality assumption by Nash~\cite{nash1953two}, i.e., the agreed solution must not be dominated.
We rewrite Eq.~\eqref{eq:nbp} as below:
\begin{equation}
    \arg\max _{\boldsymbol{\Delta}_{\boldsymbol{\theta}}} \sum_{k=1}^K \log [u_k(\Delta_{\boldsymbol{\theta}})] - \frac{\lambda}{2}  (\|\Delta_{\boldsymbol{\theta}}\|^2_2 - r).
\label{eq:lnbp}
\end{equation}
By KKT conditions~\cite{boyd2004convex},  we can get the derivative, i.e.,
\begin{equation}
\sum_{k=1}^K \frac{\Delta_{\boldsymbol{\theta}_{k}}}{{\Delta_{\boldsymbol{\theta}_{k}}^\top \Delta_{\boldsymbol{\theta}}}}  - \lambda \Delta_{\boldsymbol{\theta}} =0.
    \label{eq:dlnbp}
\end{equation}
Hence the derivative 
 of the optimal point is exactly in the radial direction, i.e., $\sum_{k=1}^K \frac{1}{{\Delta_{\boldsymbol{\theta}_{k}}^\top \Delta_{\boldsymbol{\theta}}}} \Delta_{\boldsymbol{\theta}_{k}} ||\Delta_{\boldsymbol{\theta}}$.
Considering the consistent deviations are linearly dependent, and substituting $\Delta_{\boldsymbol{\theta}}$ with Eq.~\eqref{eq:agg_grad}, we expand Eq.~\eqref{eq:dlnbp} for the inconsistent deviations with linear independent assignment, i.e., 
$\forall_k \sum_{k=1}^K p_k \Delta_{\boldsymbol{\theta}_{k}}^\top  \Delta_{\boldsymbol{\theta}_{k}} =\frac{1}{p_k}$ for $\lambda=1$.
Let $\boldsymbol{G}$ be the $d \times K$ deviation matrix whose $k-$th column is $\Delta_{\boldsymbol{\theta}_{k}}$ with dimension $d$, we obtain an equivalent, i.e., finding $\boldsymbol{p}$ in
$\boldsymbol{G}^\top \boldsymbol{G}\boldsymbol{p}=\nicefrac{1}{\boldsymbol{p}}$.
%

\nosection{Solving Nash Bargaining Problem with Approximate Multi-task Optimization}
Motivated by \cite{navon2022multi}~which efficiently approximates the optimal solution of Nash bargaining problem,  we solve $\boldsymbol{p}$ in $\boldsymbol{G}^\top \boldsymbol{G}\boldsymbol{p}=\nicefrac{1}{\boldsymbol{p}}$ through a sequence of convex optimization.
We define $q_k(p) = \Delta_{\boldsymbol{\theta}_{k}}^\top \boldsymbol{G} \boldsymbol{p}$ and seek $\boldsymbol{p}$ to solve $p_k= \nicefrac{1}{q_k}$ for all $k$, which equally shares solution with $\forall_k: \log(p_k)+\log(q_k)=0$.
We denote $\varphi_k(\boldsymbol{p}) = \log(p_k)+\log(q_k)\geq 0$ and $\varphi(\boldsymbol{p}) = \sum_k \varphi_k(\boldsymbol{p})$, and obtain:
\begin{equation}
    \begin{gathered}
\min _{\boldsymbol{p}} \varphi(\boldsymbol{p})
\quad
\text { s.t. } \forall_k:  \varphi_k(\boldsymbol{p}) \geq 0,
p_k>0,
\end{gathered}
\label{eq:eq_opt}
\end{equation}
where the constraints are convex and linear.
Under the constraints $\varphi_k(\boldsymbol{p}) \geq 0$, minimizing the convex objective $\sum_k q_k$ produces exact solutions with $\varphi(\boldsymbol{p})=0$~\cite{navon2022multi}.
Hence we introduce $\min\sum_k q_k$ to Eq.~\eqref{eq:eq_opt}, and further obtain the convex-concave approximation, i.e., 
\begin{equation}
    \begin{gathered}
\min _{\boldsymbol{p}} \Sigma_{k=1}^K q_k(\boldsymbol{p})+\varphi(\boldsymbol{p}), 
\text{ s.t. } \forall k:  \varphi_k(\boldsymbol{p}) \geq 0,
p_k>0 .
\end{gathered}
\label{eq:nash_opt}
\end{equation}

%
Expand the concave term in Eq.~\eqref{eq:nash_opt}, i.e., $\varphi(\boldsymbol{p})$, with the first-order approximation $\widetilde{\varphi}_{\tau}(\boldsymbol{p})=\varphi(\boldsymbol{p}^{\tau})+ \nabla \varphi(\boldsymbol{p}^{\tau})^\top (\boldsymbol{p}-\boldsymbol{p}^{\tau})$ for each iteration $\tau$.
As last, we obtain a convex optimization objective that can be addressed by sequential optimization~\cite{lipp2016variations}, which iteratively converges the sequence $\{\boldsymbol{p}^\tau\}_\tau$ to a critical point of the original non-convex problem in Eq.~\eqref{eq:nash_opt} by theory~\cite{lanckriet2009convergence}.
By substituting $\boldsymbol{p}$ to  Eq.~\eqref{eq:agg_grad}, we have unique Parento optimal solution for global aggregation which 
%
%
not only approaches the global optimum consistently, but also maintains the clients' model 
 optimization. 

\subsection{Overall Algorithm}
Given \moduleA~and \moduleB, we describe the overall algorithm of modeling \modelname~in Algo.~\ref{alg:fedrane}.
Steps 1:9 are the main collaboration procedure between server and clients.
Note that, each client tackles intra-client inconsistency with \moduleA~in step 5, while server handles inter-client inconsistency with \moduleB~in step 7.
Specifically, each client executes local modeling with \moduleA~to enhance representation in steps 10:20.
And server applies \moduleB~to negotiate an agreement among inconsistent client deviations, which is detailed in steps 21:25.

\begin{algorithm}[tb]
    \caption{Training procedure of \modelname}
    \label{alg:fedrane}
    \textbf{Input}: Batch size $B$, communication rounds $T$, number of clients $K$, local steps $E$, dataset $\mathcal{D} =\cup_{k\in [K]} \mathcal{D}_k$ \\
    \textbf{Output}: Global and local model parameters, i.e., $\boldsymbol{\theta}^T \text{and} \{\boldsymbol{\theta}_k^T\}^K$\\
    \begin{algorithmic}[1]
        \STATE Server initializes 
 $\boldsymbol{\theta}^{0}$
        \FOR{$t=0,1,...,T-1$}
            \FOR{$k=1,2,...,K$ \textbf{in parallel}} 
                \STATE Server sends $\{\boldsymbol{\theta}^t\}$ to client $k$
                \STATE $\boldsymbol{\theta}_k^{t+1} \leftarrow$ \moduleA\textbf{:} \textbf{Client executes}($k$, $\boldsymbol{\theta}^t$)
            \ENDFOR
        \STATE  $\boldsymbol{\theta}^{t+1} \leftarrow$ \moduleB\textbf{:} \textbf{Server executes}($\boldsymbol{\theta}^t, \boldsymbol{p}, \{\boldsymbol{\theta}^{t+1}_k\}^K$)
        \ENDFOR
        \STATE \textbf{return} $\boldsymbol{\theta}^T$ and $\{\boldsymbol{\theta}_k^T\}^K$\\
        \STATE \moduleA\textbf{:} \textbf{Client executes}($k$, $\boldsymbol{\theta}^t$)\textbf{:}
        \STATE Assign global model to the local model $\boldsymbol{\theta}_k^t \leftarrow \boldsymbol{\theta}^t$
        \FOR{each local epoch $e= 1, 2,..., E$}
            \FOR{batch of samples $(\boldsymbol{x}_{k, 1:B}, \boldsymbol{y}_{k, 1:B}) \in \mathcal{D}_{k}$}
                \STATE Feature extraction $\boldsymbol{z}_{k, 1:B} \leftarrow \mathcal{F}_{\boldsymbol{\theta}_k^e} (\boldsymbol{x}_{k, 1:B})$
                \STATE Mine the overall relational structure by Eq.~\eqref{eq:slim_l}
                \STATE Augments $\boldsymbol{z}_{k, 1:B}$ to  $\widetilde{\boldsymbol{z}}_{k, 1:B}$ by Eq.~\eqref{eq:mpn}
                \STATE Compute loss by Eq.~\eqref{eq:loss}, and update parameters of $\boldsymbol{\theta}_k^e$
            \ENDFOR
        \ENDFOR
        \STATE \textbf{return} $\boldsymbol{\theta}_k^E$ 
        \STATE \moduleB\textbf{:} \textbf{Server executes}($\boldsymbol{\theta}^t, \boldsymbol{p}, \{\boldsymbol{\theta}^{t+1}_k\}^K$)\textbf{:}
        \STATE Compute  $\{\Delta^{t+1}_{\boldsymbol{\theta}_k}\}^K$ and $\Delta_{\boldsymbol{\theta}}^{t+1}$ by Eq.~\eqref{eq:agg_grad}
        \STATE Solve for $\boldsymbol{p}$: $\boldsymbol{G}^\top \boldsymbol{G} \boldsymbol{p}= \nicefrac{1}{\boldsymbol{p}}$ by approximating Eq.~\eqref{eq:nash_opt} with sequential optimization
        \STATE Update global model with $\boldsymbol{\theta}^{t+1}= \boldsymbol{\theta}^{t} - \boldsymbol{G} \boldsymbol{p}$ 
        \STATE \textbf{return} $\boldsymbol{\theta}^{t+1}$ 
    \end{algorithmic}
\end{algorithm}

%% file: chapter/experiment.tex
\section{Experiments and Discussion}


\subsection{Experimental Setup}
\nosection{Datasets}
We conduct experiments on four benchmark datasets which are available in torchvision\footnote{https://pytorch.org/vision/stable/index.html}, i.e., EMNIST by Letters~\cite{cohen2017emnist}, Fashion-MNIST (FMNIST)~\cite{xiao2017fashion}, Cifar10, and Cifar100~\cite{krizhevsky2009learning}, following the existing \problem~work~\cite{chen2021bridging,oh2021fedbabu,li2021model}.
To evaluate \modelname, we compute both
 global performance (\textbf{G-FL}) and local personalized performance (\textbf{P-FL})~\cite{chen2021bridging}.
In detail, G-FL uses the original \textit{test set} published in the torchvision to
evaluate methods that improve the global model.
In P-FL, we compare the average local performance of methods that enhance the local models, by 
simulating non-IID local data distribution with the \textit{train set} published in torchvision.
For all datasets, we construct the non-IID data distributions via Dirichlet sampling~\cite{hsu2019measuring,li2021model}. 
That is, we sample a proportion of $j$-th class instances to client $k$ via Dirichlet distribution, i.e., $p_{j,k} \sim {Dir}_N (\alpha)$.
Smaller $\alpha$ denotes the data distributions is more heterogeneous.
We construct \textit{local training set} by randomly
sampling 75\% of local data, and  \textit{local test set} with the remaining.
%

\nosection{Comparison Methods}
We compare \modelname~with three categories of SOTA approaches by optimization goals, i.e., (1) optimizing global model: \textbf{FedAvg}~\cite{mcmahan2017communication}, \textbf{FedProx}~\cite{li2020federated}, \textbf{SCAFFOLD}~\cite{karimireddy2019scaffold}, \textbf{FedDYN}~\cite{acar2020federated}, \textbf{MOON}~\cite{li2021model}, (2) optimizing local personalized models: \textbf{FedMTL}~\cite{smith2017federated}, \textbf{FedPer}~\cite{arivazhagan2019federated}, \textbf{pFedMe}~\cite{t2020personalized},
\textbf{Ditto}~\cite{li2021ditto},
\textbf{APPLE}~\cite{luo2021adapt}, and (3) optimizing both global and local models: \textbf{Fed-RoD}~\cite{chen2021bridging}, \textbf{FedBABU}~\cite{oh2021fedbabu}, and \textbf{SphereFed}~\cite{dong2022spherefed}. 
\textbf{FedAvg} is the first vanilla federated learning framework to collaborate among server and clients.
\textbf{FedProx} takes a proximal term to regularize the change from global model to the local model.
\textbf{SCAFFOLD} considers the variance of the global model and local model when updating local gradients.
\textbf{FedDYN} applies a dynamic regularizer to pull the local model close to the global model, while pushing the local model away from  the previous local model.
\textbf{MOON} introduces contrastive learning to federated learning.
\textbf{FedMTL} is an algorithm that takes personalized learning as a multi-task learning objective.
\textbf{FedPer} captures personalization aspects in FL by decoupling neural network model and avoiding aggregating personalization layers.
\textbf{pFedMe} uses Moreau envelopes as clients’ regularized loss functions to decouple personalized model optimization from global model learning. 
\textbf{Ditto} develops a scalable solver for 
providing personalization while retaining similar efficiency. 
\textbf{APPLE} adaptively learns to personalize the client models.
\textbf{Fed-RoD}  explicitly decouples a model's dual duties with two prediction tasks. 
\textbf{FedBABU} only updates the representation body of the model during federated training, and the head is fine-tuned for personalization. 
\textbf{SphereFed} is a hyperspherical federated learning framework to 
 address \problem.
We evaluate the global model of the first and third categories of work on G-FL, and the averaged performance of local models in the second and third categories of work on P-FL.

\nosection{Implementation Details}
%
We set the number of clients $K=20$, and seek global updating direction in $B_r$ with radius $r=K$. 
We adopt ConvNet~\cite{lecun1998gradient} as the feature extractor for EMNIST and FMNIST, while ResNet~\cite{he2016deep} for Cifar10 and Cifar100.
For all of the datasets, we set batch size as 128, and embedding dimension similar to the output of the representation model, i.e., 64 for ConvNet and 512 for ResNet.
 For \modelname, we choose SGD~\cite{bonnabel2013stochastic} as the optimizer, set the learning rate $lr=0.5$, the temperature hyperparameter $\tau_1= 0.8$,
 the effect of low-rank graph $\lambda_R=0.1$, 
 and the effect of contrastive discrimination $\lambda_{\text{CD}} =0.2$.
%
%
We conduct training for all methods with 5 local epochs per round until converge.
We evaluate both G-FL and P-FL by top-1 accuracy.
We set the non-IID degree $\alpha=\{0.1, 0.5,5\}$, respectively, to test model on different degrees of heterogeneity.
%
%
%
%


\begin{figure}[t]

\centering
\subfigure[FedAvg]{
\begin{minipage}{0.47\columnwidth}{
\vspace{-0.2cm}
\centering
\includegraphics[scale=0.3]{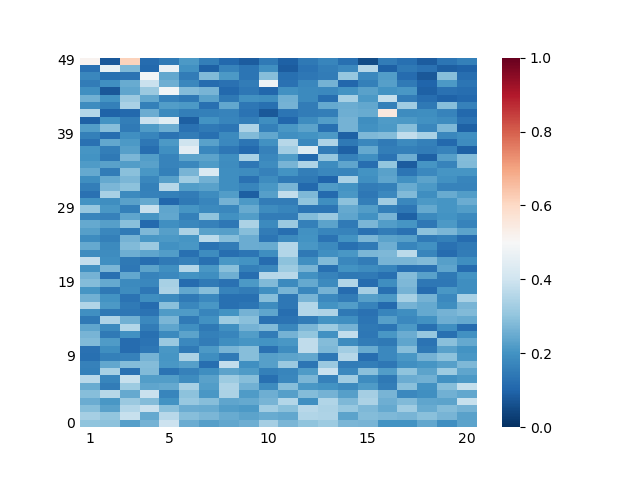} %
\vspace{-0.4cm}
}
\end{minipage}
}
\subfigure[Fed-RoD]{
\begin{minipage}{0.47\columnwidth}{
\vspace{-0.2cm}
\centering
\includegraphics[scale=0.3]{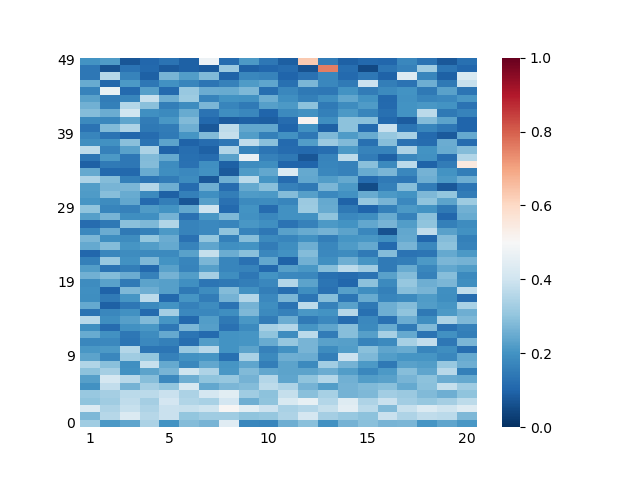} %
\vspace{-0.4cm}
}
\end{minipage}
}
\subfigure[SphereFed]{
\begin{minipage}{0.47\columnwidth}{
\vspace{-0.4cm}
\centering
\includegraphics[scale=0.3]{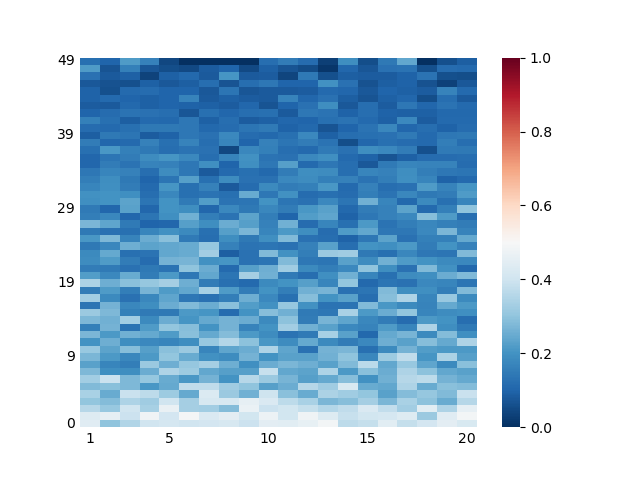} %
\vspace{-0.4cm}
}
\end{minipage}
}
\subfigure[\modelname]{
\begin{minipage}{0.47\columnwidth}{
\vspace{-0.4cm}
\centering
\includegraphics[scale=0.3]{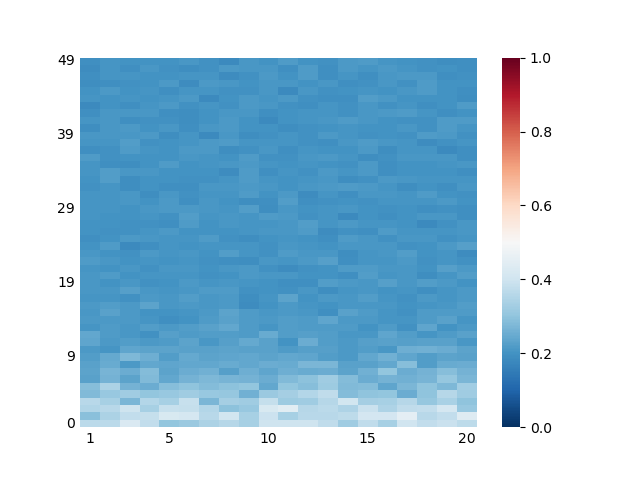} %
\vspace{-0.4cm}
}
\end{minipage}
}
\vspace{-0.1cm}
\caption{Cosine similarity between global updating direction and local deviations on Cifar10 ($\alpha=0.5$). The horizontal axis represents the client id, and the vertical axis represents the communication round. The heatmap value indicates cosine similarity and validates the model updating consistency.}
\label{fig:model_ne}
\vspace{-0.22cm}
\end{figure}

\begin{figure}[t]
\centering
\includegraphics[width=0.9\linewidth]{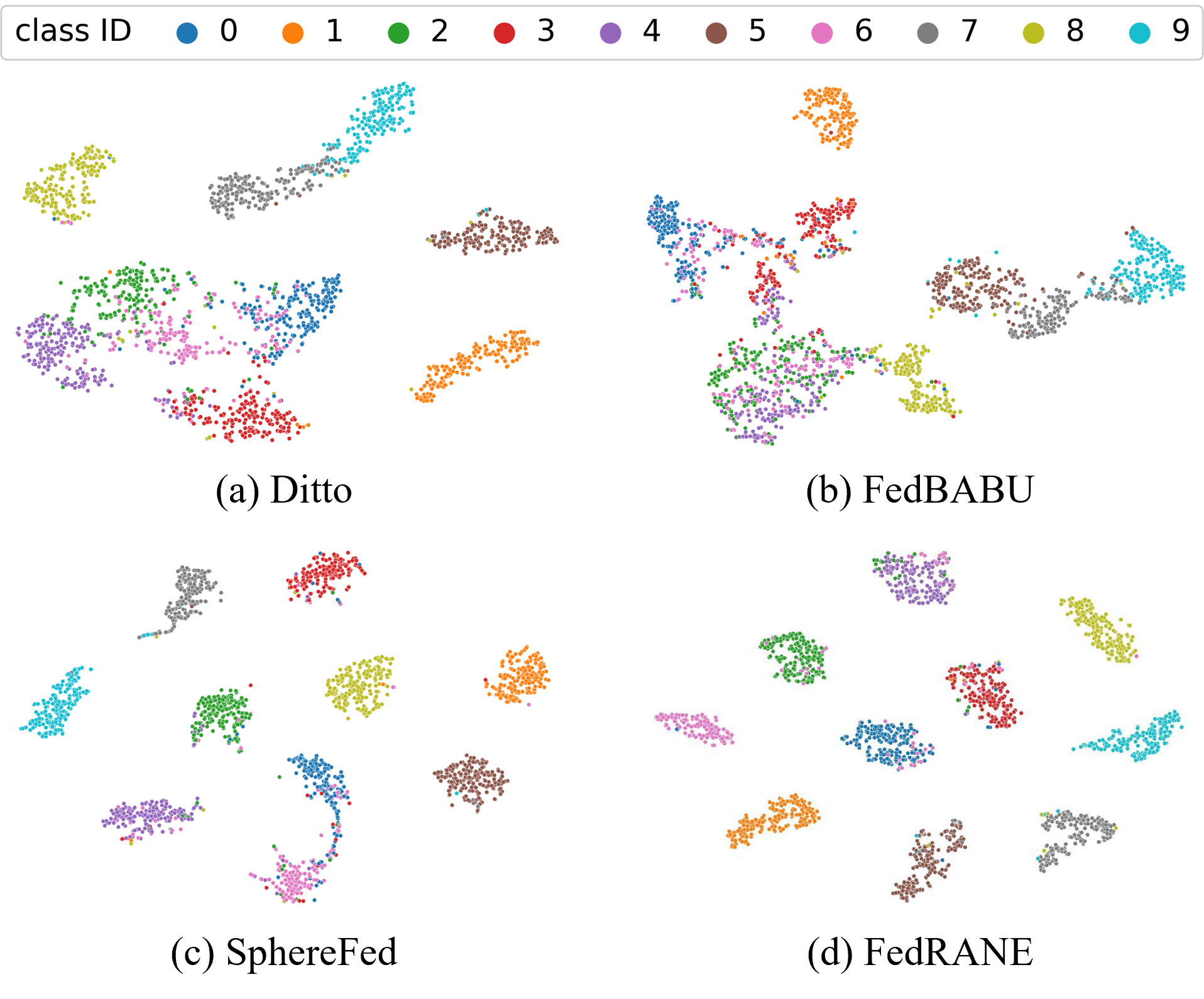}
\caption{T-SNE visualization of global representations on FMNIST ($\alpha=0.5$)}
\label{fig:model_visual}
\end{figure}

\begin{table*}[t]
\caption{Accuracy of G-FL. We bold  the best result, and underline the runner-up comparison method.}
\renewcommand\arraystretch{0.5}
\setlength\tabcolsep{6pt}
\centering
\resizebox{\textwidth}{!}{
\begin{tabular}{c|ccc|ccc|ccc|ccc}
\toprule
Dataset       & \multicolumn{3}{c}{EMNIST}   & \multicolumn{3}{c}{FMNIST}   & \multicolumn{3}{c}{Cifar10} & \multicolumn{3}{c}{Cifar100} \\
\midrule
Method \textbackslash Non-IID        & Dir(0.1) & Dir(0.5) & Dir(5) & Dir(0.1) & Dir(0.5) & Dir(5) & Dir(0.1) & Dir(0.5) & Dir(5) & Dir(0.1)  & Dir(0.5) & Dir(5) \\

\midrule
FedAvg          & 0.9011 & 0.9287 & 0.9331 & 0.7902   & 0.8685 & 0.8845 & 0.4110  & 0.6235 & 0.6758 & 0.3062 & 0.3178 & 0.3271  \\
FedProx         & 0.9010  & 0.9285 & 0.9327 & 0.7891 & 0.8678 & 0.8842 & 0.3926 & 0.6296 & 0.6726 & 0.3025 & 0.3232 & 0.3245 \\
SCAFFOLD        & 0.9077 & 0.9327 & 0.9365 & 0.7981 & 0.8747 & 0.8877 & 0.3167 & 0.6558 & 0.6993 & 0.3364 & 0.3588 & 0.3581 \\
FedDYN          & 0.9061  & 0.9257 & 0.9295 & 0.8286 & 0.8846 & 0.8972 & 0.3155 & 0.6397 & 0.6904 & 0.3209 & 0.3424 & 0.3450  \\
MOON            & 0.9028 & 0.9302 & 0.9343 & 0.8407 & 0.8966 & 0.9081  & 0.3541 & 0.5933 & 0.6393 & 0.2729 & 0.2812 & 0.3063 \\
Fed-RoD         & 0.9158 & 0.9397 & 0.9404 & 0.8421 & 0.8952 & 0.9074 & \underline{0.4434} & 0.6453 & 0.6868 & 0.3066 & 0.3332 & 0.3476 \\
FedBABU         & 0.8731 & 0.9167 & 0.9255 & 0.7591 & 0.8264 & 0.8484 & 0.3556 & 0.5966 & 0.6425 & 0.2848 & 0.3009 & 0.3080  \\
SphereFed       & \underline{0.9357} & \underline{0.9428} & \underline{0.9432} & \underline{0.8785} & \underline{0.9005} & \underline{0.9087} & 0.3393 & \underline{0.7164} & \underline{0.7488} & \underline{0.3544} & \underline{0.3781} & \underline{0.3797} \\
\midrule
\modelname-w/o-\moduleA               & 0.9388 & 0.9430  & 0.9458 & 0.8847 & 0.9092 & 0.9162 & 0.4461 & 0.7299 & 0.7494 & 0.3691 & 0.3936 & 0.4055 \\
\modelname-w/o-\moduleB                &  0.9365  & 0.9441 & 0.9465 & 0.8864 & 0.9128 & 0.9175 & 0.3861 & 0.7355 & 0.7728 & 0.3738 & 0.4110  & 0.4163 \\
\modelname                        & \textbf{0.9394} & \textbf{0.9455} & \textbf{0.9473} & \textbf{0.8892} & \textbf{0.9135} & \textbf{0.9194} & \textbf{0.5056} & \textbf{0.7407} & \textbf{0.7765} & \textbf{0.3940}  & \textbf{0.4209} & \textbf{0.4248} \\
 \bottomrule
\end{tabular}
}
\label{tb:G-FL}
\end{table*}

\begin{table*}[t]
\caption{Accuracy of P-FL. We bold  the best result, and underline the runner-up comparison method. }
\renewcommand\arraystretch{0.5}
\setlength\tabcolsep{6pt}
\centering
\resizebox{\textwidth}{!}{
\begin{tabular}{c|ccc|ccc|ccc|ccc}
\toprule
Dataset       & \multicolumn{3}{c}{EMNIST}    & \multicolumn{3}{c}{FMNIST}    & \multicolumn{3}{c}{Cifar10}  & \multicolumn{3}{c}{Cifar100} \\
\midrule
Method \textbackslash Non-IID        & Dir(0.1) & Dir(0.5) & Dir(5)  & Dir(0.1) & Dir(0.5) & Dir(5)  & Dir(0.1) & Dir(0.5) & Dir(5)  & Dir(0.1) & Dir(0.5) & Dir(5)  \\
\midrule


FedPer          & 0.9732 & 0.9373 & 0.9213 & 0.9717 & 0.9096 & 0.8755 & 0.9192 & 0.7498 & 0.6424 & 0.5227 & 0.3411 & 0.2371 \\
FedMTL          & 0.9704 & 0.9182 & 0.8855 & 0.9747 & 0.9116 & 0.8571 & 0.9012 & 0.6508 & 0.4575 & 0.4654 & 0.2638 & 0.1377 \\
pFedMe          & 0.9731  & 0.9421 & 0.9291 & 0.9611 & 0.8922 & 0.8596 & \underline{0.9262} & \underline{0.7707} & 0.6602 & \underline{0.5813} & \underline{0.4116} & 0.3313    \\
Ditto           & 0.9806 & 0.9549 & 0.9437 &\underline{0.9775} & \underline{0.9388} & \underline{0.9179} & 0.9085 & 0.7129 & 0.6292 & 0.5045 & 0.3533 & 0.2901 \\
APPLE           & 0.9740  & 0.9448 & 0.9308 & 0.9686 &  0.9074 & 0.8735 & 0.8981       &  0.6761      &  0.5613      &  0.4676      &   0.3204     &   0.2383     \\ 
Fed-RoD         & \underline{0.9831}  & \underline{0.9580}  & \underline{0.9462} & 0.9752 & 0.9359 & 0.9171  & 0.9160  & 0.7447 & 0.6906 &  0.5311 &  0.3917 & 0.3346 \\
FedBABU         & 0.9738 & 0.9415 & 0.9285 & 0.9681 & 0.8966 & 0.8566 & 0.9245 & 0.7076 & 0.6259 & 0.4734 & 0.3330  & 0.2956 \\
SphereFed       & 0.9366 & 0.9432 & 0.9454 & 0.8801   & 0.9062 & 0.9144 & 0.9121 & 0.7555 & \underline{0.7283} & 0.3496 & 0.3271 & \underline{0.3582} \\
\midrule
\modelname-w/o-\moduleA               & 0.9663 & 0.9493 & 0.9445 & 0.9662 & 0.9311 & 0.9218 & 0.9104 & 0.7588 & 0.7376 & 0.3852 & 0.3658 & 0.3931 \\
\modelname-w/o-\moduleB               & 0.9787 & 0.9562 & 0.9481 & 0.9563 & 0.9262 & 0.9264 & 0.9324 & 0.8156 & 0.7633 & 0.5636 & 0.4565 & 0.4068 \\
\modelname                        & \textbf{0.9855}   & \textbf{0.9620}   & \textbf{0.9501} & \textbf{0.9797}   & \textbf{0.9440}   & \textbf{0.9276} & \textbf{0.9347}   & \textbf{0.8270 }   & \textbf{0.7687} & \textbf{0.6144}    & \textbf{0.4701}   & \textbf{0.4162}\\
\bottomrule
\end{tabular}
}
\label{tb:P-FL}
\end{table*}

\begin{figure}[t]

\centering
\subfigure[G-FL]{
\begin{minipage}{0.47\columnwidth}{
\vspace{-0.2cm}
\centering
\label{fig:cifar10_nc_GFL}
\includegraphics[scale=0.48]{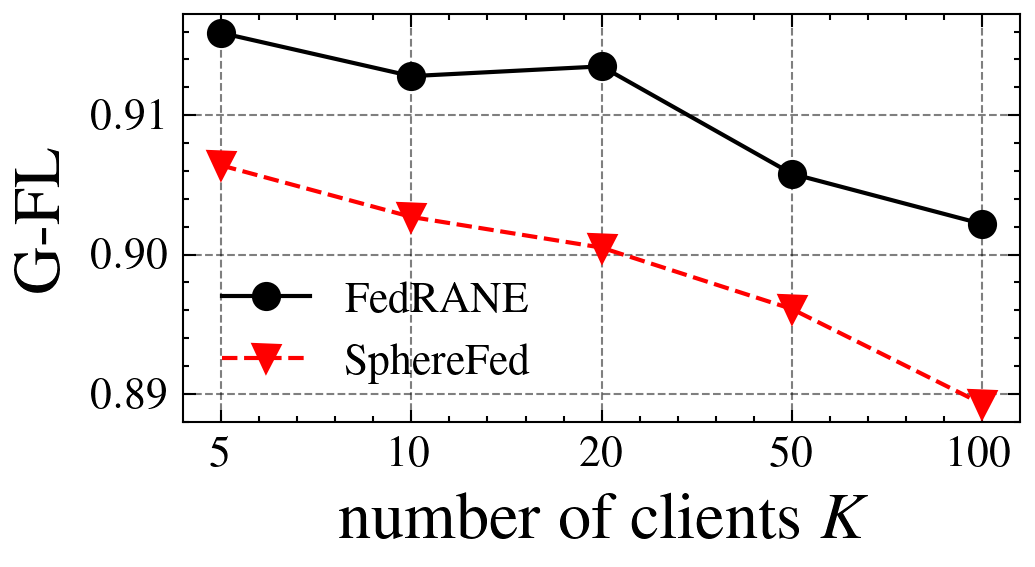} %
\vspace{-0.4cm}
}
\end{minipage}
}
\subfigure[P-FL]{
\begin{minipage}{0.47\columnwidth}{
\vspace{-0.2cm}
\centering
\label{fig:cifar10_nc_PFL}
\includegraphics[scale=0.48]{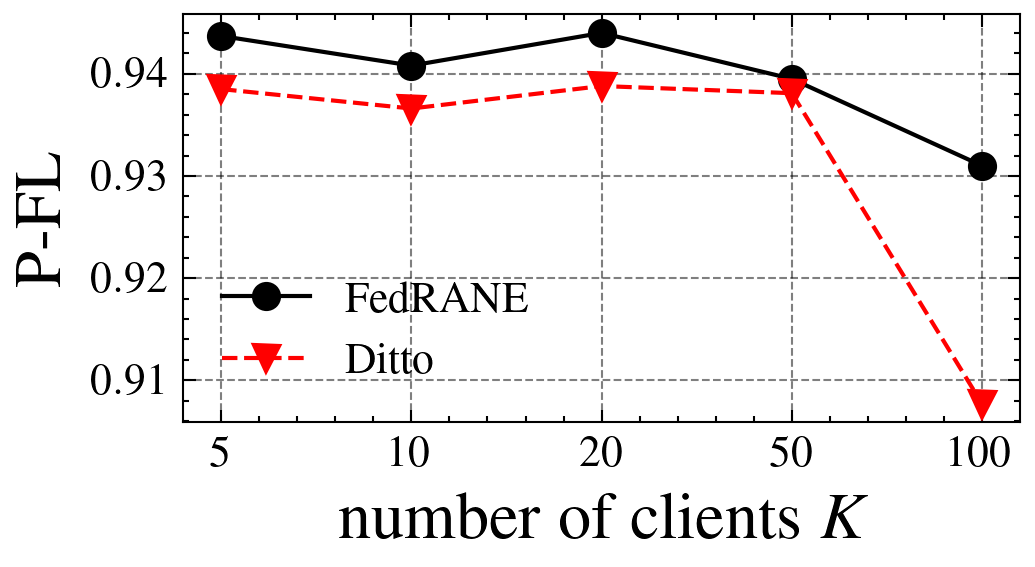} %
\vspace{-0.4cm}
}
\end{minipage}
}
\vspace{-0.1cm}
\caption{Effect of the client numbers $K$ on FMNIST ($\alpha=0.5$)}
\label{fig:n_clients}
\end{figure}



\begin{figure}[t]
\centering
\subfigure[G-FL]{
\begin{minipage}{0.46\columnwidth}{
\vspace{-0.1cm}
\centering
\label{fig:cifar10_E_gfl}
\includegraphics[scale=0.47]{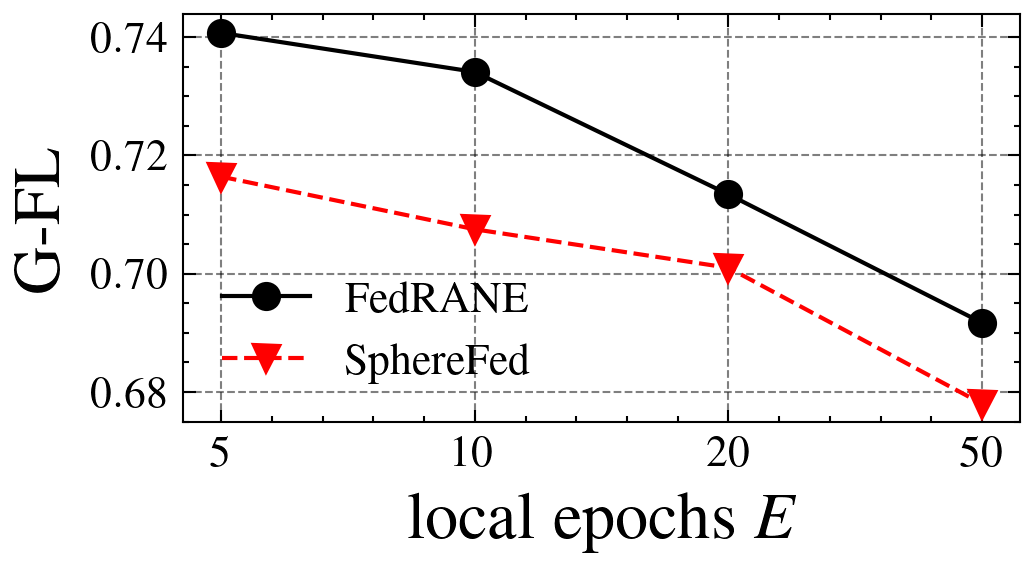} %
\vspace{-0.4cm}
}
\end{minipage}
}
\subfigure[P-FL]{
\begin{minipage}{0.46\columnwidth}{
\vspace{-0.2cm}

\centering
\label{fig:cifar10_E_pfl}
\includegraphics[scale=0.47]{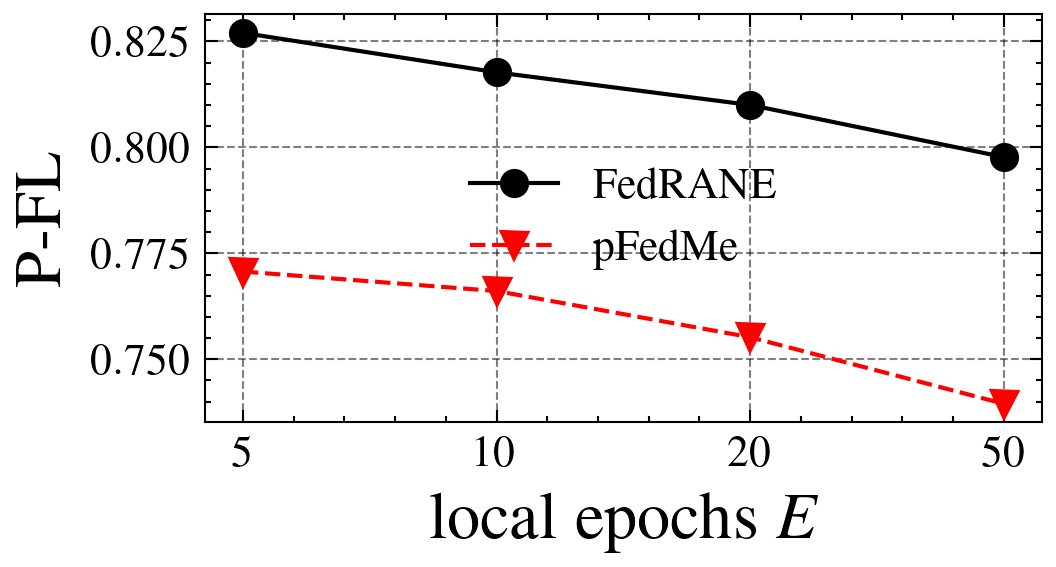} %
\vspace{-0.4cm}
}
\end{minipage}
}
\vspace{-0.1cm}
\caption{Effect of Local Epochs $E$ on Cifar10 ($\alpha=0.5$)}
\label{fig:hyperparameter_E}
\end{figure}

\begin{figure}[t]
\vspace{-0.24cm}
\centering
\subfigure[G-FL]{
\begin{minipage}{0.47\columnwidth}{
\centering
\label{fig:cifar10_cd_reg_GFL}
\includegraphics[scale=0.48]{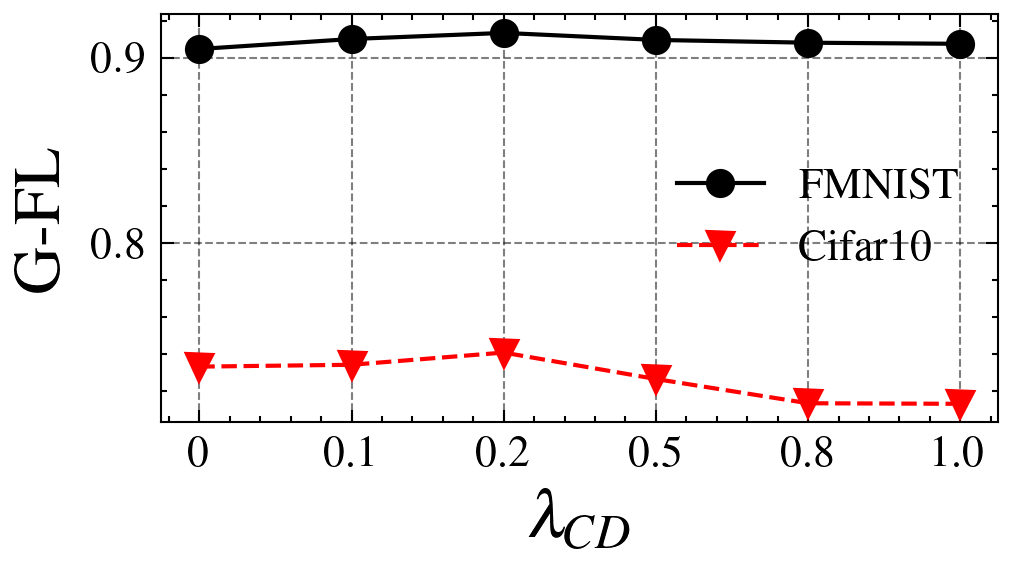} %
\vspace{-0.4cm}
}
\end{minipage}
}
\subfigure[P-FL]{
\begin{minipage}{0.47\columnwidth}{
\centering
\label{fig:cifar10_cd_reg_PFL}
\includegraphics[scale=0.48]{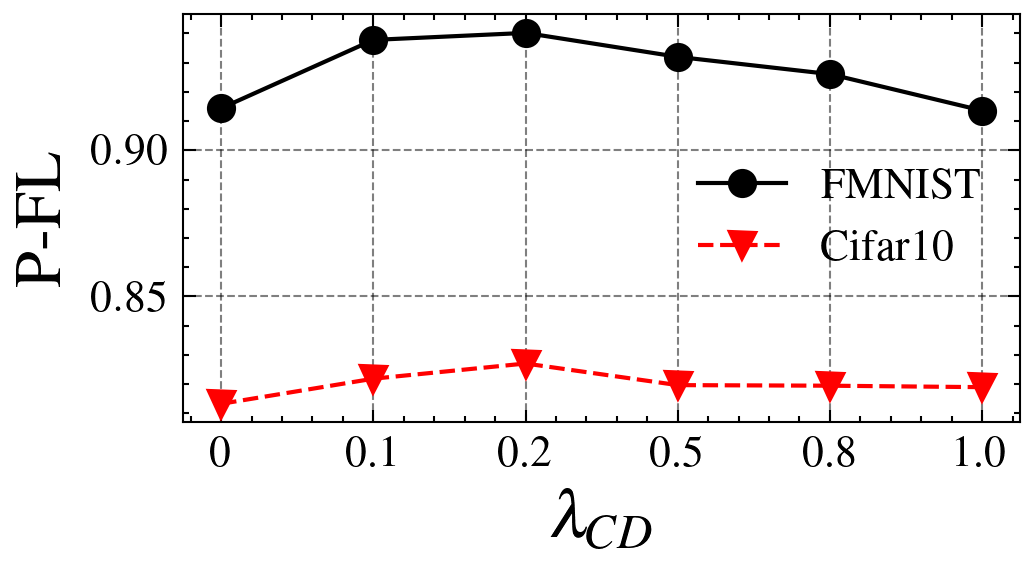} %
\vspace{-0.4cm}
}
\end{minipage}
}
\vspace{-0.1cm}
\caption{Effect of $\lambda_{\text{CD}}$ on FMNIST and Cifar10 ($\alpha=0.5$)}
\label{fig:reg_cd}
\end{figure}

\begin{figure}[t]
\vspace{-0.24cm}
\centering
\subfigure[G-FL]{
\begin{minipage}{0.46\columnwidth}{
\centering
\label{fig:cifar10_tau_GFL}
\includegraphics[scale=0.47]{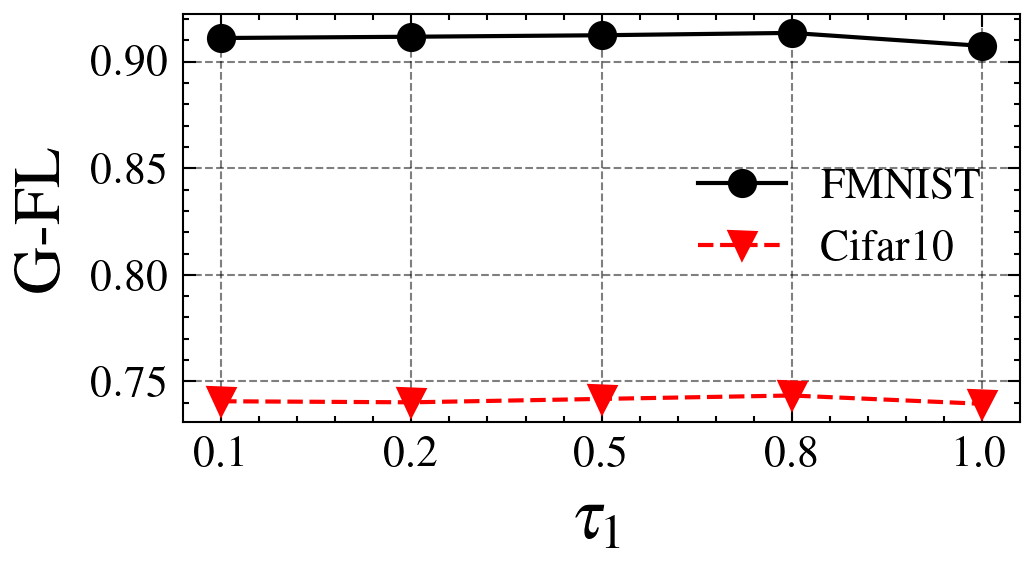} %
\vspace{-0.4cm}
}
\end{minipage}
}
\subfigure[P-FL]{
\begin{minipage}{0.46\columnwidth}{
\centering
\label{fig:cifar10_tau_PFL}
\includegraphics[scale=0.47]{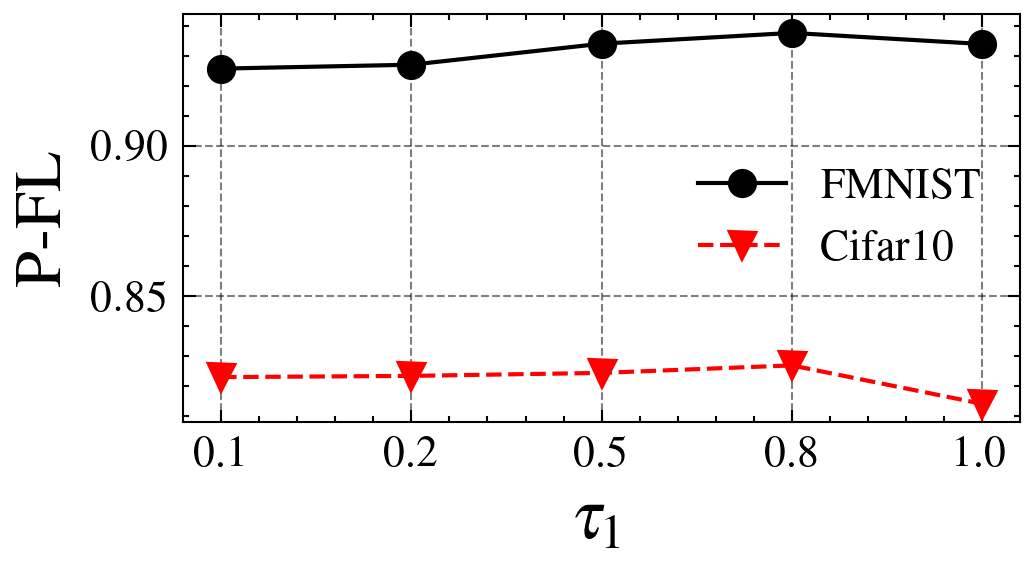} %
\vspace{-0.4cm}
}
\end{minipage}
}
\vspace{-0.1cm}
\caption{Effect of temperature $\tau_1$ on FMNIST and Cifar10 ($\alpha=0.5$)}
\label{fig:temperature}
\end{figure}

\subsection{Empirical Results}
\nosection{Performance Comparison}
For every method, we conduct the experiments with its best parameters five times and report the average value for both G-FL and P-FL in Tab.~\ref{tb:G-FL}-\ref{tb:P-FL}, respectively.
We get the conclusions based on three main observations.
(1) \textbf{In terms of G-FL} evaluated in Tab.~\ref{tb:G-FL}, 
the larger degree of non-IID, i.e., a smaller $\alpha$ in $\operatorname{Dir}(\cdot)$, challenges more on all methods.
Though G-FL methods achieve satisfying performance on simple tasks, i.e., EMNIST and FMNIST, they get degradation seriously on tough tasks, especially on Cifar10 and Cifar100 $(\alpha=0.1)$. 
The third category of methods mainly outperforms the first category of methods, rectifying that decoupling the impact of global and local optimization can improve the global model.
(2) \textbf{In terms of P-FL} evaluated in Tab.~\ref{tb:P-FL},
the performance of P-FL methods decreases with the increase of $\alpha$, meaning that the personalization performance relies on the data portion of the same class in each client.
In other words, P-FL methods achieve better results when $\alpha=0.1$ due to the fact that each client accounts heavily for the majority of samples.
The severe degradation of the second and third categories when $\alpha=\{0.5, 5\}$ implies that the current P-FL methods fail to capture the representations of the minority data samples well.
%
(3) \textbf{According to the performance of \modelname} in both Tab.~\ref{tb:G-FL} and Tab.~\ref{tb:P-FL}, \modelname~outperform most of methods, with the advantage of tackling intra- and inter-client inconsistencies simultaneously.
Compared with the runner-up method in G-FL, 
the performance improvement in smaller $\alpha$ is  generally larger on tough tasks, 
i.e., Cifar10 and Cifar100.
This states that with the updating agreement of inconsistent deviations, \modelname~not only obtains the better global optimization to global optimum, but also keeps the local optimization towards local optimum unchanged.
Compared with the runner-up method in P-FL, 
\modelname~decreases performance less, since \moduleA~refines the sufficient representations for the minority.

\nosection{Visualization}
We compare cosine similarity between global updating direction and local deviations in Fig.~\ref{fig:model_ne}, to validate the consistency between the global updating and client model deviations.
%
We can find \modelname~updates global model with a direction that is consistent and balanced among all client model deviations.
This brings both better global and local model performance as stated in Tab.~\ref{tb:G-FL} and Tab.~\ref{tb:P-FL}.
Besides, we sample 2,000 samples and visualize their feature representations of the global model using t-SNE~\cite{van2008visualizing} in Fig.~\ref{fig:model_visual}.
Note that, \modelname~obtains more separable decision bound via tackling both intra- and inter-client inconsistencies.

\nosection{Ablation Studies}
%
%
We study the effectiveness of \moduleA~and~\moduleB~via two variants of \modelname: 
(1) \modelname~without applying \moduleA, i.e., \modelname-w/o-\moduleA,
and (2) \modelname~substituting \moduleB~with average weighted by sample ratio, i.e., \modelname-w/o-\moduleB.
Firstly, from Tab.~\ref{tb:G-FL} and Tab.~\ref{tb:P-FL}, we can find that both \modelname-w/o-\moduleA, and \modelname-w/o-\moduleB~mainly degrade their performance compared with \modelname.
This validates that handling inconsistency simultaneously will obtain the superior performance.
%
Secondly, note that \modelname-w/o-\moduleA, \modelname-w/o-\moduleB, and \modelname~achieve slightly similar performance on EMNIST and FMNIST, this means tackling either intra- or inter-client inconsistencies improves the G-FL performance on simple tasks.
Lastly, both \moduleA~and \moduleB~contribute to addressing \problem.
Compared with the runner-up method, \modelname-w/o-\moduleB~still obtains better performance in P-FL with larger $\alpha$, meaning that \moduleA~corrects the modeling of imbalanced data in serious non-IID.
%
\modelname-w/o-\moduleA~is better than the runner-up in G-FL,
via negotiating an agreement for inconsistent client deviations improves both local and global performance.

\nosection{Hyper-parameters sensitivity}
We study the sensitivity of highly relevant                                                 hyper-parameters on FMNIST and Cifar10 ($\alpha=0.5)$.
We tune the number of clients $K= \{5, 10, 20,50,100\}$ in Fig.~\ref{fig:n_clients}, the local epochs $E= \{5, 10,20, 50\}$ in Fig.~\ref{fig:hyperparameter_E}, the effect of contrastive discrimination $\lambda_{\text{CD}}=\{0, 0.1, 0.5,0.8 1\}$ in Fig.~\ref{fig:reg_cd}, and the effect of temperature in contrastive discrimination $\tau_1=\{0.1,0.2, 0.5,0.8, 1\}$ in Fig.~\ref{fig:temperature}, respectively.
From the accuracy curves, we can conclude:
%
(1) The performance of all methods decreases when the number of clients increases, but \modelname~can perform better than the runner-ups, i.e., SphereFed in G-FL, and Ditto in P-FL.
%
(2) With the increase of local epochs, the model deviations among clients will increase, making it harder to obtain a well-performed FL model.
\modelname~maintain its effectiveness stably, since \moduleB~can handle the inter-client inconsistency and obtain Pareto optimal solution for both global and local performance.
(3) The two hyper-parameters of contrastive discrimination, i.e., $\lambda_{\text{CD}}$ and $\tau_1$, slightly impact the performance of G-FL, but change the performance of P-FL evidently.
%
